\documentclass[runningheads]{llncs}
\usepackage{graphicx}
\usepackage{amsmath}
\usepackage{booktabs}

\newcommand{\pa}{\textbf{point-adjust}~}
\newcommand{\pw}{\textbf{point-wise}~}
\newcommand{\ew}{\textbf{event-wise}~}
\begin{document}
\title{Multivariate Time Series Anomaly Detection: Fancy Algorithms and Flawed Evaluation Methodology}
\titlerunning{~}
\author{Mohamed El Amine Sehili \and Zonghua Zhang}
\authorrunning{Sehili and Zhang}
\institute{Huawei Paris Reasearch Center, 8 Quai du Point du Jour, 92100 Boulogne-Billancourt, France}
\maketitle              % typeset the header of the contribution

\begin{abstract}
Multivariate Time Series (MVTS) anomaly detection is a long-standing and challenging research
topic that has attracted tremendous research effort from both industry
and academia recently. However, a careful study of the literature makes
us realize that 1) the community is active but not as organized as other sibling
machine learning communities such as Computer Vision (CV) and Natural Language
Processing (NLP), and 2) most proposed solutions are evaluated using either
inappropriate or highly flawed protocols, with an apparent lack of scientific
foundation. So flawed is one very popular protocol, the so-called \pa protocol,
that a random guess can be shown to systematically outperform \emph{all} algorithms
developed so far. In this paper, we review and evaluate many recent algorithms
using more robust protocols and discuss how a normally good protocol may have
weaknesses in the context of MVTS anomaly detection and how to mitigate them.
We also share our concerns about benchmark datasets, experiment design and evaluation
methodology we observe in many works. Furthermore, we propose a simple, yet challenging,
baseline based on Principal Components Analysis (PCA) that surprisingly outperforms many
recent Deep Learning (DL) based approaches on popular benchmark datasets. The main
objective of this work is to stimulate more effort towards important aspects
of the research such as data, experiment design, evaluation methodology and result
interpretability, instead of putting the highest weight on the design of increasingly
more complex and ``fancier'' algorithms\footnote{code associated with this
paper is at https://github.com/amsehili/MVTSEvalPaper}.

\keywords{Multivariate time series  \and Anomaly detection \and Evaluation protocols \and point-adjust.}
\end{abstract}

\section{Introduction: MVTS Anomaly Detection, a Hot Research Topic}
\label{sec:intro}
%--------------------------------------------------------------------------------

Time series are a kind of data characterized by its ease of collection and storage as well
as its wide range of applications such as forecasting, classification and anomaly detection.

Anomaly detection in multivariate time series, particularly unsupervised one, is a very
important topic for the modern, data-powered, industry. This explains the high interest
in the subject in both industry and academia and the proliferation of approaches
in recent years, especially deep learning (DL) based ones. While one should salute 
the agility and effectiveness of the community at leveraging advancements from other
fields such as NLP and CV, and repurposing them for anomaly detection, one may also argue
that evaluation methodology, benchmark datasets and objective algorithms' comparison are
still open challenges in the field as of today \cite{kim2022towards,wu2021current}.

Our point is straightforward: algorithm design and pipeline complexity have been granted
much more effort than experiment design and methodology. Actually, over the last decade
or so, works on MVTS anomaly detection have adapted and experimented with a few of the most
recent and most influential ideas in deep learning such as Generative Adversarial Networks
(GANs) \cite{audibert2020usad,chen2021daemon}, Transformers \cite{xu2021anomaly} and
Graph Neural Networks (GNNs) \cite{deng2021graph,zhang22grelen,han2022learning}. At the
same time, we notice that a significant number of approaches are evaluated in terms of one
very flawed protocol, the \pa protocol, making it impossible, as we demonstrate in this paper,
to know whether an algorithm is outputting random noise as a prediction or doing something cleverer.
This may sound overly pessimistic, but as formally proven by \cite{kim2022towards}, a random
guess-based approach can outperform all ``sophisticated'' algorithms on popular datasets when
evaluated with this protocol.

In this work, we review the \pa protocol and show that by randomly selecting a small, fixed,
number of points and tagging them as anomalous, while considering the rest of the points as normal,
we can achieve very high scores with high probabilities. Our goal is to
continue warning the community against this protocol and its derivatives and encourage it to drop
it in favor of more reliable protocols. We also review the more objective \pw protocol and
show that it is not appropriate for all kinds of datasets and use-cases. We then introduce
and discuss the \ew protocol, a new \emph{event-aware} protocol for MVTS anomaly detection.

Besides, we demonstrate that an approach as basic as Principal Components Analysis (PCA),
with simple pre-processing and post-processing blocks, can outperform many complex
DL-based approaches on many datasets. Finally, we analyze three of the most
recent approaches and show how algorithms developed by exclusively targeting
high \pa scores fail to distinguish themselves from a random guess when challenged
with other protocols. Finally, without claiming to be an authority in the field, we
share what, we believe, are vectors of improvement based on our experience working on
the topic for many years at an industrial level.

This paper is organized as follows. In section \ref{sec:relatedwork} we review related work.
Section \ref{sec:evalproto} discusses issues of popular evaluation protocols and introduces
alternative, more robust protocols for MVTS anomaly detection. Section \ref{sec:dscaveats} highlights
a few issues related to benchmark datasets and experiment design we frequently observe in
the literature. In section \ref{sec:algos} we evaluate and compare a few of the most recent
algorithms using many protocols. Finally, in section \ref{sec:discussion}, we outline
what we consider to be better practices towards a more objective evaluation of MVTS anomaly
detection approaches.

\section{Related Work}
\label{sec:relatedwork}

%--------------------------------------------------------------------------------

While the number of works on MVTS anomaly detection has been increasing over the
years, very few studies about issues in benchmark datasets, algorithms and especially
evaluation protocols and metrics have been proposed so far.

In their paper entitled ``Current Time Series Anomaly Detection Benchmarks are
Flawed and are Creating the Illusion of Progress'' \cite{wu2021current}, Wu and Keogh
argue that many popular univariate and multivariate benchmark datasets for anomaly
detection have serious flaws that make the claimed performance of many algorithms
questionable. The authors demonstrate that using the so-called \emph{one-liners}
(i.e., very short code with a simplistic logic such as the difference between two
consecutive points or the moving average), one can achieve state-of-the-art performance
on the datasets considered in the study. They particularly pinpoint four flaws that
the majority of the studied datasets suffer from: triviality, unrealistic anomaly
density, mislabeled ground truth and run to failure bias (i.e., anomalies systematically
located at the end of time series).

While we agree with most of these points and find the conclusions of the paper highly
important, we believe that the point about the unrealistic anomaly density in datasets
should not always be regarded as a flaw. Actually, as the authors point out, anomalies
in data are inherently rare, and it is difficult to build a real-world benchmark dataset
that contains $10\%$ or $15\%$ of anomalies, for example. However, having many anomalies
in the data, ideally with different distributions, is necessary to reliably evaluate
algorithms and their generalization capability. Many of the frequently used MVTS benchmark
datasets (e.g., SWaT \cite{mathur2016swat}, Wadi \cite{ahmed2017wadi} and PSM
\cite{abdulaal2021practical}) are gathered from real-world industrial systems and contain
a relatively high ratio of anomalies obtained by deliberately altering the normal functioning
of the system. These datasets are good to evaluate unsupervised anomaly detection approaches
because their training part is anomaly free\footnote{Assumption based on the description shared
by datasets' publishers. In any case, \emph{no anomaly labels} are provided for the training
fold of these datasets, and almost all approaches using them are unsupervised.}.

Nevertheless, we believe that this point may become concerning when \emph{training} data are
provided with an ``abnormally'' high ratio of \emph{labeled} anomalies, making it possible
to use supervised approaches. This may be problematic because in the real-world, supervised
algorithms would not have access to such big amounts of labeled anomalies for training.

In terms of evaluation protocols and metrics, recent works on MVTS anomaly detection have
been marked by the use of the highly defective \pa protocol (discussed in section
\ref{sec:point-adjust}), claiming very high scores. Kim et al. \cite{kim2022towards} propose
the first thorough review of this protocol, exposing its flaws. They particularly introduce
three basic methods that outperform all state-of-the-art algorithms evaluated using this
protocol. These methods include: 1) drawing a random anomaly score from a uniform distribution
over $[0, 1]$, 2) using raw feature values as an anomaly score, and 3) predicting anomalies
using an \emph{untrained} neural network.

Based on the conclusions of \cite{kim2022towards},  Garg et al. \cite{garg2021evaluation}
propose the \textbf{composite} protocol in which the recall is calculated based on the number
of detected anomalous \emph{events} (as opposed to anomalous points, see section
\ref{sec:alt-proto}), whereas the precision is calculated, as usual, at the point level. The
F1 score is then computed using the \emph{event-level} recall and \emph{point-level} precision.
This protocol alleviates the effects of lengthy events on the \pw protocol (introduced in
section \ref{sec:point-wise}) and allows for a more intuitive evaluation of algorithms. With
such a protocol, we can, for example, know how many events an algorithm has detected. This is
important information that the \pw protocol does not provide.

\section{Evaluation Protocols for MVTS Anomaly Detection}
\label{sec:evalproto}

\subsection{Point-adjust: a Non-protocol for Time Series Anomaly Detection}
\label{sec:point-adjust}

\begin{figure}
\centering
\includegraphics[scale=0.85]{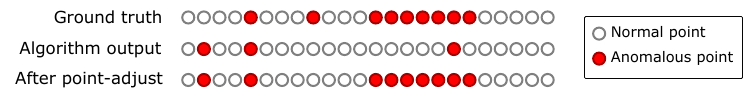}
\caption{Illustration of the \pa evaluation protocol. A single anomalous point
detected by the algorithm implies that all other points within the segment be 
counted as True Positive, even if the algorithm did not actually detect them.}
\label{fig:paproto}
\end{figure}

According to this protocol, if an algorithm correctly detects at least one
anomalous point within a segment of many, contiguous, anomalous points, then all
points within the segment are systematically considered detected and count as
True Positive (TP). The rationale behind this is that one segment represents 
one anomalous event (which is a quite reasonable assumption) and thus, if the
algorithm detects one point within it as anomalous, then: 1) the whole event
should be declared as successfully detected, and 2) the algorithm should be
\emph{rewarded} with as many TP as there are points in the segment, as illustrated
in Fig. \ref{fig:paproto}.

While the fact of ``successfully detecting one anomalous point within a segment
is sufficient to consider the event detected'' is straightforward, we find the
second part of the reasoning problematic. First, it starts off by presenting
the evaluation criterion as event-based: the most important is whether the event
is detected once, regardless of the number of actually detected anomalous points
within it. The reasoning then seamlessly switches to a point-wise evaluation,
multiplying, \emph{on its way there}, the number of TP by the \emph{length} of
reasoning would be to count just \emph{one} TP for \emph{one} detected event, like
in protocols described in section \ref{sec:alt-proto}.

Second, this reasoning seems to rely on the tacit assumption that anomalous points
within the same segment are \emph{similar}, and therefore it is ``acceptable''
to detect just one or a few of them to consider all of them successfully detected.
We believe that, if this assumption holds, then from an algorithm developer's
perspective, the fact that the algorithm only detects a subset of presumably similar
anomalous points should be an argument to investigate the behavior of the
algorithm instead of rewarding it with points it did not detect. If the assumption
does not hold, however, and anomalous points within the same segment may have different
distributions, then considering that all the points were correctly detected based on
the mere fact that one of them was detected by the algorithm sounds deliberately
misleading.

As a result, using this protocol, an algorithm that has many false alarms but happens
to detect one or a couple of anomalies within an anomalous segment would achieve a high
score. Based on the distribution of anomalies in many frequently used benchmark datasets,
a very efficient such an algorithm would be a one that \emph{randomly} selects a small
number of points and tag them as anomalous while considering the remaining points as
non-anomalous. This section introduces, analyses and evaluates an algorithm of this kind.

Kim et al. \cite{kim2022towards} show that this protocol can be \emph{hacked} using
several procedures, including sampling random anomaly scores from a uniform distribution,
using raw feature values as an anomaly score, and predicting anomalies using an \emph{untrained}
neural network. They also provide a formal proof for the first case.

%--------------------------------------------------------------------------------
In this work, we propose a procedure to reach a \pa F1 score of choice, and
calculate the probability of reaching it. The goal of the current part is two-fold: 1)
show how manipulable the \pa protocol can be, and 2) lay the ground for discussing
the weaknesses of the F1 score in general and how inappropriate it can be for some
kinds of datasets and anomaly distributions. The latter will be further discussed
in section \ref{sec:point-wise}.

The F1 score is calculated in terms of precision $\text{P}$ and recall $\text{R}$ as follows:

\begin{equation*}
    \text{F1} = \frac{2 \times \text{P} \times \text{R}}{\text{P} + \text{R}} \quad \text{for} \quad \text{P} = \frac{\text{TP}}{\text{TP} + \text{FP}} \quad \text{and}  \quad \text{R} = \frac{\text{TP}}{\text{TP} + \text{FN}}
\end{equation*}

%--------------------------------------------------------------------------------

\noindent where TP stands for True Positive, FP for False Positive, and FN for False
Negative.

We define $r$, the contamination rate in evaluation data, that is, the ratio of
anomalous points to the total number of points. If we randomly pick up $\alpha$
points from evaluation data and tag them as anomalous, then the probability of
\emph{not} hitting any anomalous point is $(1 - r)^{\alpha}$. Thus, the probability
of hitting one anomalous point at least is $1 - (1 - r)^{\alpha}$.

Based on this, if the anomalies make up one single segment, then applying the \pa
procedure yields a perfect recall, $\text{R}_{pa}=1$ with the same probability:

\begin{equation}
\label{eq:pa_R1}
p(\text{R}_{pa} = 1 | r, \alpha) = 1 - (1 - r)^{\alpha}
\end{equation}

This is the probability of selecting at least one single point within the anomalous
segment with $\alpha$ trials, which is also the probability of selecting \emph{at most}
($\alpha - 1$) points from outside the segment. In other words, this is the probability
of having ($\alpha - 1$) false alarms at most: $p(\text{FP}\le\alpha - 1 | r, \alpha)$.

The \pa precision, $\text{P}_{pa}$, obtained from adjusted TP, $\text{TP}_{pa}$, is:

\begin{equation}
\text{P}_{pa} = \frac{\text{TP}_{pa}}{\text{TP}_{pa} + \text{FP}}
\end{equation}

Let $A$ be the length of the anomalous segment. When $\text{R}_{pa} = 1$, then
$\text{TP}_{pa} = A$ and FP equals $(\alpha - 1)$ at most. Thus, the worst
\pa precision is $\text{P}_{pa} = \frac{A}{A + (\alpha - 1)}$. We can write
the probability of having a \pa precision of at least $\frac{A}{A + (\alpha - 1)}$,
given $r$ and $\alpha$, as:

\begin{equation}
\label{eq:p_Ppa}
p(\text{P}_{pa} \ge \frac{A}{A + (\alpha - 1)} |r, \alpha) = 1 - (1 - r)^{\alpha}
\end{equation}

Similarly, when $\text{R}_{pa} = 1$ we can write $\text{F1}_{pa}$ as:

\begin{equation}
\label{eq:f1parpa1}
\text{F1}_{pa} = \frac{2 \times \text{P}_{pa} \times 1}{\text{P}_{pa} + 1} = \frac{2 \times \text{P}_{pa}}{\text{P}_{pa} + 1}
\end{equation}

From Eq. \ref{eq:p_Ppa} and Eq. \ref{eq:f1parpa1} we have:

\begin{equation*}
p(\text{F1}_{pa} \ge \frac{2 \times \frac{A}{A + (\alpha - 1)}}{\frac{A}{A + (\alpha - 1)} + 1} |r, \alpha) = 1 - (1 - r)^{\alpha}
\end{equation*}

\noindent which can be written as:

\begin{equation}
\label{eq:pf1parpa1}
p(\text{F1}_{pa} \ge \frac{2A}{2A + \alpha - 1} | r, \alpha) = 1 - (1 - r)^{\alpha}
\end{equation}

Eq. \ref{eq:pf1parpa1} means that the larger $\alpha$, the more confident we are about the
\emph{minimum} $\text{F1}_{pa}$ we can achieve by tagging $\alpha$ random points as anomalous,
but the lower is that \emph{minimum} value itself. However, we can also see that larger values of
$A$ yield higher values for the \emph{minimum} $\text{F1}_{pa}$ value because
$\lim_{A\to\infty} \left[\frac{2A}{2A + \alpha - 1}\right] = 1$.

\begin{figure}
\centering
\includegraphics[width=\textwidth]{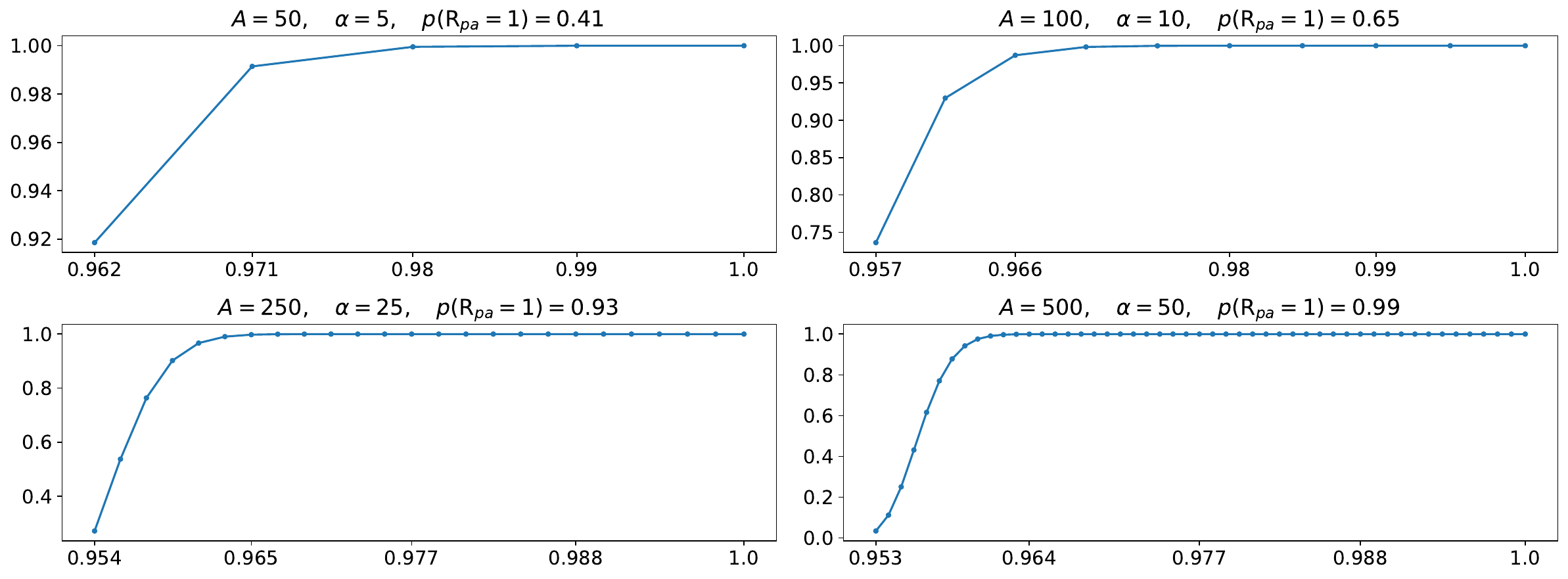}
\caption{Probability of achieving a perfect \pa recall ($p(\text{R}_{pa} = 1$) for
four different datasets that have the \emph{same} contamination rate, $r=0.1$, but
different anomalous segment's lengths. For each dataset, we randomly select $1\%$ of
the
points ($\alpha$) and tag them as anomalous. Each subplot represents the CDF of
the corresponding $\text{F1}_{pa}$ score. That is, the x-axis represents threshold
values and the y-axis is the probability of obtaining an $\text{F1}_{pa}$ score $\le$
the corresponding threshold value. Each value corresponds to $s \in [1, \alpha]$, the
number of successes hitting the anomalous segment. We can see that for the same
contamination rate across datasets, we are more confident about obtaining fairly
high $\text{F1}_{pa}$ scores as the length of the anomalous segment increases.
For $A=500$ for example, the probability of having $\text{F1}_{pa} \le 0.953$
is close to $0$.}
\label{fig:f1pacdf}
\end{figure}

To illustrate this, we consider four hypothetical datasets with different sizes but the
\emph{same} contamination rate, $r=0.1$, as shown in Fig. \ref{fig:f1pacdf}\footnote{These
values are not arbitrary but are close to what we observe in popular benchmark datasets,
especially for a $A=500$.}. Each dataset contains one anomalous segment whose length is
$10\%$ of the total dataset length. For each dataset, we compute the probability of having
a perfect \pa recall ($p(\text{R}_{pa} = 1)$) by randomly tagging $1\%$ of the points as
anomalous, and show the corresponding Cumulative Distribution Function (CDF) of the 
$\text{F1}_{pa}$ score, that is, the probability of $\text{F1}_{pa}$ being $\le$ a given
value.

\begin{figure}
\centering
\includegraphics[width=\textwidth]{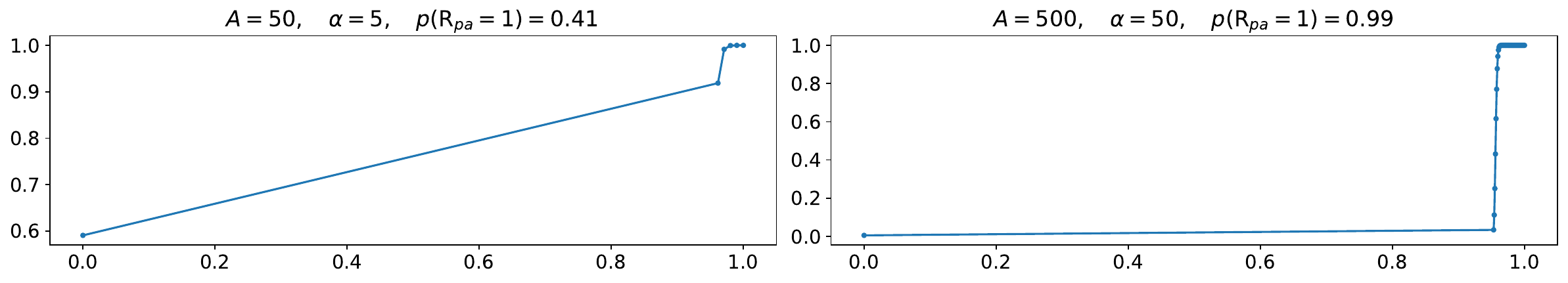}
\caption{Probability of achieving a perfect \pa recall ($p(\text{R}_{pa} = 1$)
for $A=50$ and $A=500$. Unlike Fig. \ref{fig:f1pacdf}, this figure shows the
CDF starting from $\text{F1}_{pa}=0$.}
\label{fig:f1pacdffull}
\end{figure}

As can be observed in Fig. \ref{fig:f1pacdf}, the larger $A$, the higher $p(\text{R}_{pa} = 1)$
and the less likely that $\text{F1}_{pa}$ falls below the fairly high value of about
$0.95$. Actually, for $A=50$, we have $p(\text{F1}_{pa}=0) = 0.59$ and for $A=500$, $p(\text{F1}_{pa}=0) = 0.005$.
For illustration purposes, these values are not shown on Fig. \ref{fig:f1pacdf} but
in a separate figure, Fig. \ref{fig:f1pacdffull}. Fig. \ref{fig:f1pacdffull} is
similar to Fig. \ref{fig:f1pacdf} (for $A=50$ and $A=500$) but shows the CDF of
$\text{F1}_{pa}$ starting from $\text{F1}_{pa}=0$. We can see that with the \pa
procedure, detecting one single anomaly within an anomalous segment may result in
an unrealistically big jump in $\text{F1}_{pa}$.

%--------------------------------------------------------------------------------

\begin{figure}
\centering
\includegraphics[scale=0.5]{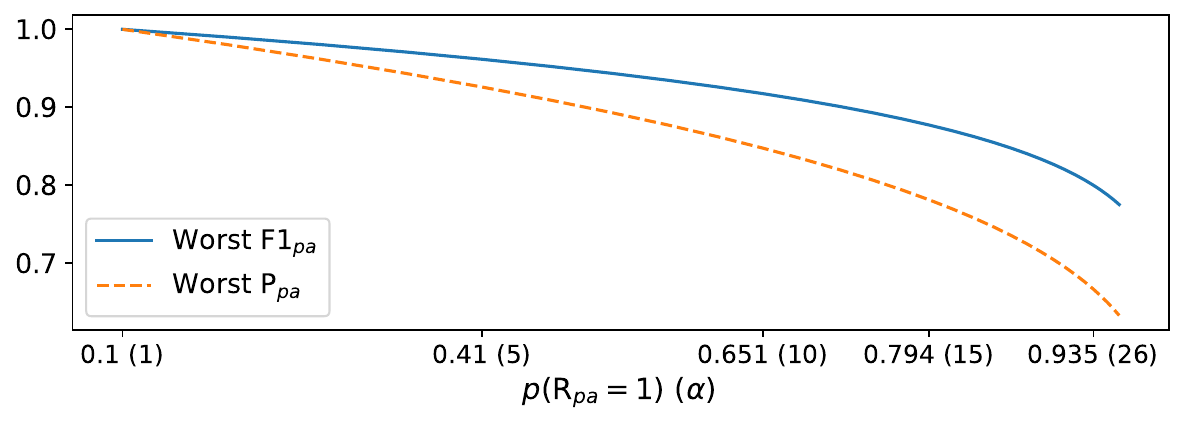}
\caption{Probability of achieving a perfect \pa recall ($p(\text{R}_{pa} = 1$) for $A=50$ and
$r=0.1$ for different values of $\alpha$, as well the \emph{worst} corresponding values of
$\text{F1}_{pa}$ and $\text{P}_{pa}$. The \emph{worst} values correspond to the case where
one single point within the anomalous segment and $\alpha - 1$ points from outside the
segment are respectively tagged as anomalous. Each $x$ value corresponds to $p(\text{R}_{pa}=1)$
given a value of $\alpha$.}
\label{fig:f1paworst}
\end{figure}

In Fig. \ref{fig:f1pacdf} and \ref{fig:f1pacdffull}, the probability of achieving a perfect
\pa recall for $A=50$ is $0.41$ and that of having $\text{F1}_{pa}=0$ is fairly high ($0.61$).
These probabilities are based on $\alpha = 5$. It turns out that we can improve these scores
and their probabilities by using \emph{better} values for  $\alpha$ while keeping other parameters
unchanged (i.e., $A=50$ and $r=0.1$). Fig. \ref{fig:f1paworst} shows the probability of achieving
a perfect \pa recall for different values of $\alpha$, and the worst resulting $\text{P}_{pa}$
and $\text{F1}_{pa}$ scores. These worst scores are reached if, out of $\alpha$ randomly selected
points, one single point falls within the anomalous segment and $\alpha - 1$ points outside of it.
For $\alpha=26$, for example, $p(\text{F1}_{pa} \ge 0.8) \approx 0.935$.

In practice, using this procedure with $\alpha = 1000$ yielded an average $\text{F1}_{pa}$ score
of about $0.95$ for SWaT and Wadi datasets and $0.98$ for PSM. These scores are higher than the
ones obtained using elaborate DL-based pipelines.

\subsection{Point-wise: a Good Protocol but not for All Situations}
\label{sec:point-wise}
This protocol consists of computing the precision, recall and F1 score based on the
actual output of the algorithm, without any ``adjustment''. There is nothing special
about this procedure, which is used in many other fields. One interesting feature of the
F1 score is that it is generally a good default choice for unbalanced datasets.
An unbalanced dataset contains a dominant class with the majority of points belonging
to it\footnote{Machine learning students are usually advised to use F1 as a better
alternative to the \emph{accuracy} score for unbalanced datasets because using the latter
would yield a high score for a trivial algorithm that predicts the dominant class for
every input.}. However, in many cases, including anomaly detection, it is often desirable
to know the number of false alarms raised by an algorithm or at least their rate, referred
to as False Alarm Rate (FAR), and computed as the number of false alarms to the total
number of normal points.

Consider for instance an anomaly detection algorithm used to detect upcoming hard
disk failures in a data center. Detecting an upcoming failure means that the disk should
be replaced. Using the FAR as one of the metrics to evaluate the algorithm helps 
estimate the expected number of unduly replaced disks and the resulting cost. This is
not possible when using the precision or the F1 score.

\begin{figure}
\centering
\includegraphics[scale=0.45]{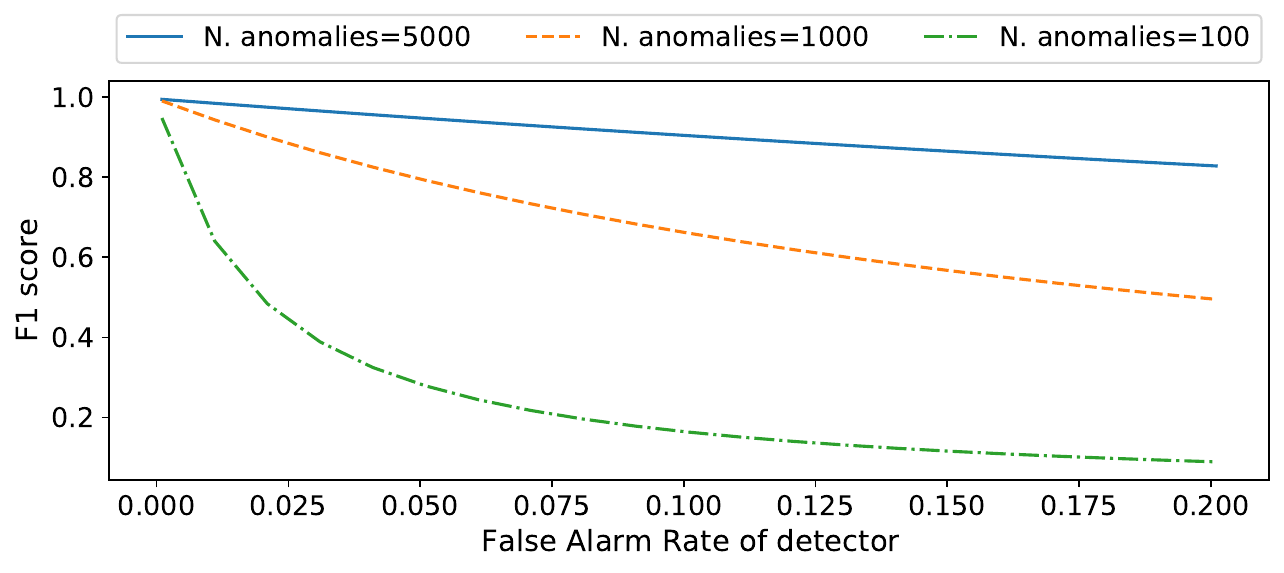}
\caption{F1 score of many anomaly detectors on three different datasets. All detectors have
the same recall of 0.99, but a different False Alarm Rate (i.e., each detector has
its own FAR). Each curve corresponds to a dataset that has $10000$ normal points
but a different number of anomalous points (the datasets have, therefore, a different 
contamination rate). We can see that, even if the recall and the FAR of a detector are
the same for all datasets, its F1 score may be much higher for datasets with a high
contamination rate than for datasets with a low contamination rate. This is particularly
true for detectors with a high FAR.
}
\label{fig:detvarpos}
\end{figure}

To further illustrate the limits of the F1 score, we consider many anomaly detectors
that have all the same recall of $0.99$, but a FAR that varies across detectors
($0.001 \le \text{FAR} \le {0.2}$). We compute the F1 score of each of these detectors
with three datasets that have the \emph{same} number of normal points ($10000$) but
a different number of anomalous points (5000, 1000 and 100 points respectively, meaning
that the datasets have a different contamination rate). Fig. \ref{fig:detvarpos}
shows the F1 score of each of these detectors for the three datasets. As can be observed,
for a given detector (one tick on the x-axis), the F1 scores vary considerably depending
on the number of anomalous points in the dataset.

This happens because the precision and, as a result, the F1 score, are also a function of
TP: a higher TP results in a better precision and F1 score. In fact, as the FAR of a detector
is the same across datasets, and as the three datasets have the same number of normal points,
the expected FP is the same for the detector across datasets. However, as the number of anomalous
points is not the same for all datasets, the TP of a detector is higher for datasets with more
anomalies. As a result, the precision and F1 score are better for datasets with a higher
contamination rate. Consequently, the precision and the F1 score can only yield a \emph{coarse}
estimation of false alarms and can hardly be used to accurately estimate their cost in terms
of money, time, human effort, etc.

More related to time series, another case where F1 score computed in the \pw fashion has limits
is when anomalies make up lengthy segments rather than isolated outliers. We refer to these
segments as anomalous \emph{events}. In principle, if two algorithms have roughly the same FAR,
one would prefer the one that detects \emph{more} events even if its \pw recall is lower. The
\pw F1 score does not allow algorithm comparison from that perspective, and does not let us
know how many events were ever detected by an algorithm.

The limits of the \pw F1 score may be exacerbated for benchmark datasets that contain one or a
couple of very long events compared to the rest of events. In such cases, an algorithm may be tuned
to efficiently detect anomalies within the lengthy events, while probably (and seamlessly)
being less efficient for shorter events\footnote{This situation has subtle links to the one of
imbalanced datasets for which the F1 score is usually recommended in the first place.}. This is
for example the case for the SWaT dataset, in which the median event length is 450 points, but
there is one 35K-point event that is, furthermore, the most \emph{anomalous} and the easiest
to detect.

\subsection{Alternative Evaluation Protocols}
\label{sec:alt-proto}

\subsubsection{The Composite Protocol}
By design, the \pw F1 score treats TP and FP alike (and also FN, but this point in not
necessary to the current discussion). As a result, each TP is \emph{celebrated} as one
more detected anomaly and each FP is regarded as a false alarm. The problem is that,
for benchmark datasets that contain anomalous events, the very first anomaly detected within
an event is normally more informative than consecutively detected anomalies within the same event.
Hence, the symmetry with which \pw protocol considers TP and FP is no longer justified.
Based on this reasoning, Garg et al. \cite{garg2021evaluation} propose to count just one TP for each
detected event, regardless of the number of anomalous points detected within it. The recall
is hence computed at the \emph{event level}. However, to compute the
precision, the TP and FP at the \emph{point level} are used, just like with the \pw protocol. The
resulting F1 score, referred to as $\text{F1}_{C}$ in the following, is called \textbf{composite}.
This protocol is much more appropriate than the \pw protocol for datasets in which anomalies
take the form of long events, and we believe it should be paid more attention in future works.

\subsubsection{The Event-wise Protocol}
Motivated by the conclusions of \cite{garg2021evaluation}, and in an attempt to 1) partially
disentangle the precision from TP and give more weight to false alarms, and 2) make the performance
of algorithms easier and more intuitive to interpret\footnote{By interpretation we mean understandinghow many
anomalous events the algorithm detects and how many false alarm it raises. This has nothing to do
with model \emph{interpretability} whose goal is to answer \emph{why/how} an algorithm made a
given decision.}, we propose the \ew protocol.

The \ew protocol computes TP at the event level, as in the \textbf{composite} protocol. However,
FP is not counted at the point level, as with the \pw and \textbf{composite} protocols, but is the
number of anomalous \emph{segments} found by the algorithm that do not overlap with \emph{any}
Ground Truth (GT) event. For clarity, we use ``segment'' to refer to any set of contiguous points
that the algorithm considers anomalous, and ``event'' to refer to a GT anomalous event. Based on
this, we define the following \ew metrics:

\begin{itemize}
    \item $\text{TP}_{E}$: number of GT events that fully or partially overlap with one segment or more.
        The \ew recall is calculated as $\text{R}_{E} = \frac{\text{TP}_{E}}{\text{\# GT events}}$. This
        is the same recall used in the \textbf{composite} protocol.
    \item $\text{FN}_{E}$: number of GT events that do not overlap with any segment.
    \item $\text{FP}_{E}$: number of segments that do not overlap with any GT event. This includes segments
    of many points as well as isolated points. The \ew precision can be computed as
    $\text{P}_{E} = \frac{\text{TP}_{E}}{\text{TP}_{E} + \text{FP}_{E}}$.
\end{itemize}

\begin{figure}
    \centering
    \includegraphics[scale=0.775]{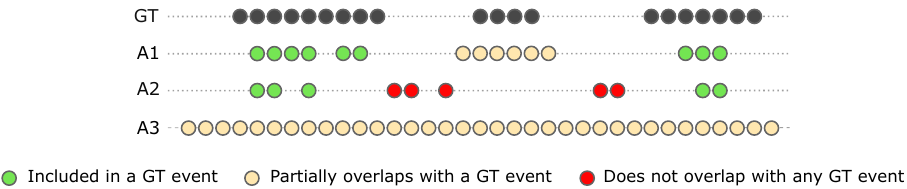}
    \caption{Illustration of the \ew protocol with three GT events and the outputs of three algorithms.
    Algorithm A1 detects all events without false alarms (i.e., it has no segment that does not overlap
    with any event). Despite this, its precision is not perfect because its second segment is too long.
    Algorithm A2 achieves an \ew recall of 2/3 and a low precision because its $\text{FP}_{E}=3$.
    Algorithm A3 detects all events but its \ew precision, $\text{P}_{E}$, and its $\text{F1}_{E}$ score
    equal $0$ because its $\text{FAR}$ is $1$.}
    \label{fig:ew-proto-exp}
\end{figure}

A similar protocol was proposed by \cite{hundman2018detecting} but has not been much used in consecutive
works. This is likely due to one serious issue it has: an algorithm that predicts an alarm for every point
(e.g., algorithm A3 in Fig. \ref{fig:ew-proto-exp}) would have prefect \ew recall ($\text{R}_{E}$) and
precision ($\text{P}_{E}$). To avoid this, we propose to involve the FAR at the point level
in the computation of the \ew precision. The precision at the event level is therefore computed as:

\begin{equation}
    \text{P}_{E} = \frac{\text{TP}_{E}}{\text{TP}_{E} + \text{FP}_{E}} \times (1 - \text{FAR})
\end{equation}

\noindent where $\text{FAR} = \frac{\text{FP}}{N}$ and $N$ is the number of normal points in data. The \ew F1 score,
denoted $\text{F1}_{E}$, is then computed in terms of $\text{R}_{E}$ and $\text{P}_{E}$.

Fig. \ref{fig:ew-proto-exp}
illustrates this protocol with the outputs of three algorithms. Note that algorithm A3 has $\text{F1}_{E}=0$
even though it has a \emph{perfect} \ew recall. This is the case because this algorithm
has a $\text{FAR}=1$. Thanks to the introduction of $\text{FAR}$ in the computation of the precision,
the proposed \ew protocol is more sensitive to false alarms compared to other protocols, and is less dependent
on TP or on the number of anomalous points in the dataset.

\section{Benchmark Datasets, Experiment Design and Algorithms Comparison}
\label{sec:dscaveats}
Wu and Keogh \cite{wu2021current} discuss flaws of a few datasets for univariate and
multivariate time series used for anomaly detection, and how trivial solutions can achieve
state-of-the-art performance on them. In this section, we discuss other issues related
to a few popular MVTS datasets and certain practices in terms of evaluation
that hinder objective algorithm comparison.

SWaT and Wadi are two very popular MVTS benchmark datasets for anomaly
detection. They are fairly big in terms of number of points and have each many anomalous
events introduced by deliberately manipulating parts of the system from which they are
collected. However, these datasets are shared by their publishers in a raw format upon
subscription, with incoherent labeling for the former and two data versions for the latter.

The SWaT dataset, for example, has one integrated label column as well as a separate
file that contains the start and the end of anomalous events. Using events' start and end
to reconstruct the labels does not result in the same content as in the integrated label
column. Wadi has two versions from 2017 and 2019, with different sizes and different
anomaly ratios. This situation has led to a disagreement about the characteristics
of these datasets across papers.

Furthermore, as mentioned earlier, the SWaT dataset contains one anomalous event
that is much longer than other events. This event contributes to the F1
score much more than other events and explains the relatively high scores of
many algorithms using the \pw protocol with this dataset compared to other datasets.
Actually, by shortening this event to the median event length, \cite{garg2021evaluation}
showed a significant drop in the \pw F1 score for all algorithms. While this
is an interesting finding, we do not advocate for deliberate event shrinking
because it may introduce artifacts and subtle changes in the data, making the event
\emph{easier} to detect. We believe that this issue can be better
dealt with using event-aware evaluation protocols like the ones introduced in
section \ref{sec:alt-proto}

The use of different evaluation protocols across publications has also led to
situations where authors evaluate their algorithm with the \pa protocol but
report results of other approaches using the \pw protocol. This is for example
observed in \cite{chen2021learning}, which evaluates the proposed algorithm
using \pa and compares the results to at least one algorithm (GDN
\cite{deng2021graph}) using the \pw protocol. Other works, such as
\cite{tuli2022tranad}, use the \pa protocol without even mentioning it
in the text. These practices, which may be observed in other works (and which we
assume are not intentional), are very likely to be misleading for uninformed readers,
including reviewers. Finally, in \cite{pan2022duma} the authors explicitly confirm
that they do \emph{not} use the \pa protocol, citing what is probably the
main work highlighting the flaws of this protocol as of today, \cite{kim2022towards}.
However, reported scores for the proposed algorithm, as well as for state-of-the-art
algorithms considered for comparison, look as high as the usual \pa scores encountered
in many studies with the same datasets.
\section{Algorithms}
\label{sec:algos}

In this part, we evaluate three DL-based algorithms for MVTS anomaly detection,
as well as a baseline algorithm based on PCA. These DL-based approaches are selected
because they are quite recent, introduce very interesting ideas, and
achieve a high performance with the used protocols. Moreover, all
of these approaches have an official open-source implementation that we
used in our experiments. The approaches are:

\begin{itemize}
    \item {\bfseries Anomaly Transformer} \cite{xu2021anomaly}, uses Transformers
      to model time series dynamic patterns. Evaluated using the \pa protocol
      only in original work.
    \item {\bfseries NCAD} \cite{carmona2021neural}, uses a 1D Convolutional Neural
      Network (CNN) to obtain feature representations of data. Also evaluated using
      the \pa protocol only in original work.
    \item {\bfseries GDN} \cite{deng2021graph}, uses GNNs to learn the relationship
      between time series to achieve a better forecasting performance. Evaluated
      using the \pw protocol only in original work.
\end{itemize}

Our goal is to challenge these algorithms with different evaluation protocols and
confront them with PCA. Table \ref{tab:res-pw-comp-ew} summarizes the obtained
results with \textbf{point-wise}, \textbf{composite} and \ew protocols.
It also reports the number of detected events ($\text{TP}_{E}$) and the number
of predicted anomalous segments that do not overlap with any GT event ($\text{FP}_{E}$,
see Fig. \ref{fig:ew-proto-exp}). Results are obtained using the SWaT (with labels
constructed from attacks' start and end), Wadi (2017 version), and PSM datasets.

\begin{table*}[h!]
    \centering
{\renewcommand{\arraystretch}{1}
    \resizebox{\textwidth}{!}{%
    \begin{tabular}{rcccccccccccc}
    \toprule
                     & \multicolumn{4}{c}{\bfseries SWaT}     & \multicolumn{4}{c}{\bfseries Wadi} & \multicolumn{4}{c}{\bfseries PSM} \\
    \cmidrule(rl){2-5} \cmidrule(rl){6-9} \cmidrule(rl){10-13}

       & \multicolumn{1}{c}{F1} & \multicolumn{1}{c}{$\text{F1}_{C}$} & \multicolumn{1}{c}{$\text{F1}_{E}$} & \multicolumn{1}{c}{$\text{TP}_{E}$/$\text{FP}_{E}$} & \multicolumn{1}{c}{F1} & \multicolumn{1}{c}{$\text{F1}_{C}$} & \multicolumn{1}{c}{$\text{F1}_{E}$} &  \multicolumn{1}{c}{$\text{TP}_{E}$/$\text{FP}_{E}$} & \multicolumn{1}{c}{F1} & \multicolumn{1}{c}{$\text{F1}_{C}$} & \multicolumn{1}{c}{$\text{F1}_{E}$} & \multicolumn{1}{c}{$\text{TP}_{E}$/$\text{FP}_{E}$}  \\

     \cmidrule(rl){2-5} \cmidrule(rl){6-9} \cmidrule(rl){10-13}

    \textbf{AT}   & ~0.214 & ~0.214 & ~0.000 & 35/0   & ~0.108 & ~0.108 & ~0.000 & 14/0    & ~0.434  & ~0.434 & ~0.000 & 72/0 \\
    \midrule

    \textbf{NCAD} & ~0.217 & ~0.217 & ~0.002 & 35/694 & ~0.114 & ~0.115 & ~0.003 & 14/1394 & ~0.429  & ~0.429 & ~0.000 & 72/0 \\
    \midrule

    \textbf{GDN~} & ~0.821 & ~0.488 & ~0.478 & 11/0   & ~0.567 & ~0.764 & ~0.485 & 9/14    & ~0.594 & ~0.640  & ~0.096 & 63/843 \\
    \midrule
    
    \textbf{PCA~} & ~0.810 & ~0.596 & ~0.555 & 15/4   & ~0.374 & ~0.655 & ~0.608 & 7/2     & ~0.538 & ~0.484  & ~0.200 & 32/170 \\ 

    \bottomrule
    \\
    \end{tabular}

   }
}
\caption{Comparative results of AnomalyTransformer (AT), NCAD, GDN and PCA  using
the \pw ($\text{F1}$), \textbf{composite} ($\text{F1}_{C}$) and \ew ($\text{F1}_{E}$)
protocols. All metrics are computed based on the detection threshold that yields
the best \pw performance. The SWaT, Wadi and PSM datasets have 35, 14 and 72 events 
and a contamination rate of $11.98\%$, $5.71\%$ and $27.76\%$ respectively.}
\label{tab:res-pw-comp-ew}
\end{table*}

These results reveal several interesting findings, summarized as follows:

\begin{itemize}
    \item Algorithms that were developed using \pa as the sole target fail
    to reach any score better than a random guess when evaluated with other
    protocols. This is the case of \textbf{AnomalyTransformer} and \textbf{NCAD}.
    For completeness, we also evaluated these algorithms using the
    \pa protocol and achieved the same, very high, scores reported in the original papers.
    It is worth mentioning that we also achieved essentially the same high scores using
    \emph{untrained} versions of these models. We assume that many other 
    approaches developed using the same setting would face the same problem\footnote{
    Also check out this issue and related ones on AnomalyTransformer's official
    repository: https://github.com/thuml/Anomaly-Transformer/issues/34
    }.

    \item \textbf{GDN}, however, which was developed based on the
    more realistic \pw protocol, shows more resilience when evaluated with
    other protocols.

    \item Datasets that have a very high contamination rate, such as PSM, yield
    \pw F1 scores that can be misleading. \textbf{AnomalyTransformer} and
    \textbf{NCAD}, for example, achieve a \pw F1 score around $0.43$, which may
    look as a \emph{fairly good} baseline score. In fact, based on the contamination
    rate of the dataset, this score is achievable by predicting all or most
    points as anomalous, and this is what these algorithms are doing.
    The \textbf{composite} score is comparable to the \pw score in this case 
    and does not bring any useful information. The \ew score, however, severely
    penalizes such approaches thanks to the inclusion of FAR in score computation.

    \item Finally, we show that PCA, which is not considered as a particularly
    advanced approach for MVTS anomaly detection, achieves an honorable \pw score
    compared to many recent DL-based approaches (e.g.,
    USAD \cite{audibert2020usad}, MSCRED \cite{zhang2019deep}, OmniAnomaly
    \cite{su2019robust} and DAGMM \cite{zong2018deep}, see \cite{kim2022towards}
    for a summary of the performance of these approaches).  Our conclusion here
    is that many works have been developed without establishing a simple but
    enough challenging baseline. Actually, the bar of $0.8$ \pw F1 score for the
    SWaT dataset had been a symbolic target for many years until recent GNN-based
    approaches, such GDN \cite{deng2021graph} and FuSAGNet \cite{han2022learning},
    reached it. In our PCA-based pipeline, we use simple pre-processing and
    post-processing blocks (input scaling, clipping and score smoothing) that
    significantly improve the score.
\end{itemize}

\begin{figure}
    \centering
    \includegraphics[scale=0.4]{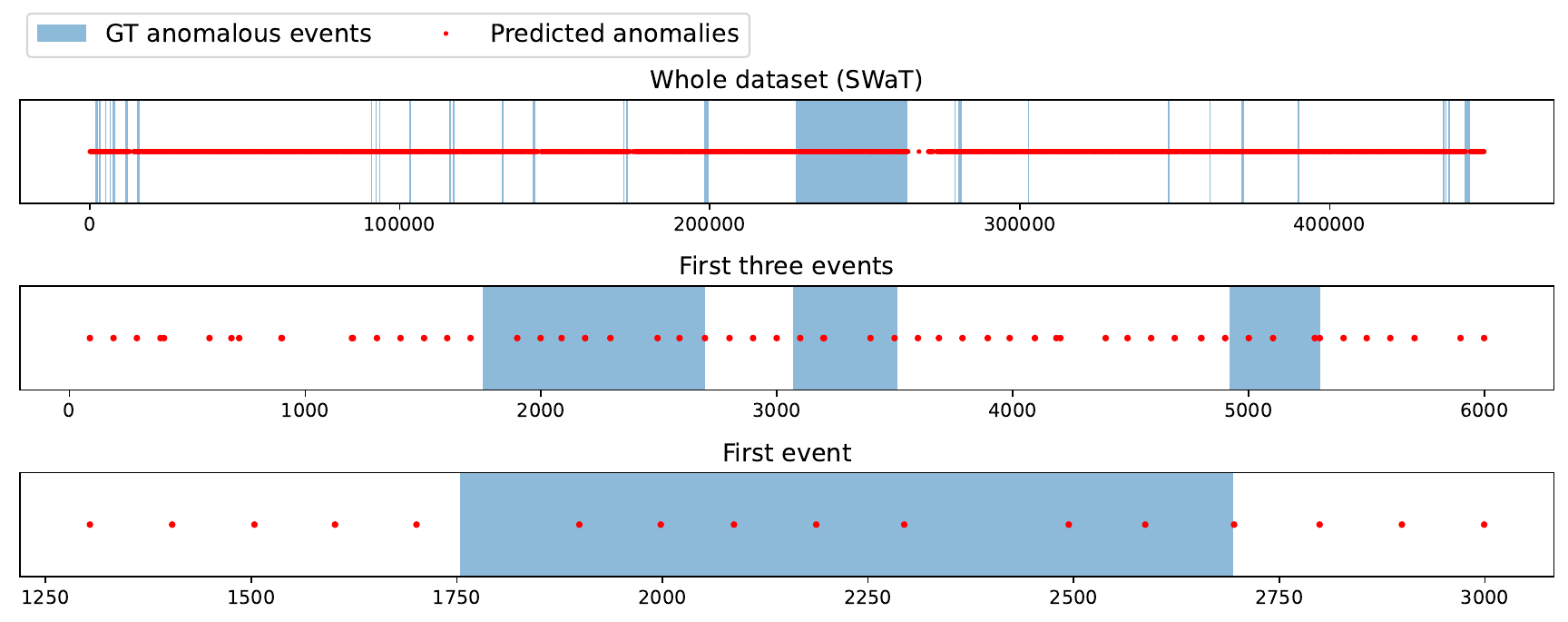}
    \caption{Ground Truth anomalous events of the SWaT dataset and outputs
    of \textbf{AnomalyTransformer}. The second and third rows zoom in on the
    first three events and on the first event, respectively. We can see that all the
    algorithm is doing is output an anomaly prediction at almost regular
    intervals. This ensures a $0.97$ $\text{F1}_{pa}$ score. Note that on the
    first row (whole dataset), there are exactly 4097 red dots ($\le1\%$
    of the total points) but they look many more because the size of the dots
    is intentionally big for visualization purposes.}
    \label{fig:swat-gt-anotrans-pred}
\end{figure}

Scores in Table \ref{tab:res-pw-comp-ew} are obtained using the threshold that
yields the best \pw F1 score. To further understand the decisions of
algorithms such as \textbf{AnomalyTransformer} and \textbf{NCAD}, we used
the threshold that yields the best \pa score for these algorithms and looked at their
outputs to better understand their behavior. As aforementioned, using the \pa protocol,
we obtained scores comparable to those reported 
in original papers for these algorithms (actually, we achieved even higher scores).
The first row in Fig. \ref{fig:swat-gt-anotrans-pred} shows all GT events of SWaT
as well as the outputs of \textbf{AnomalyTransformer} that lead to a $0.97$ $\text{F1}_{pa}$
score. By zooming in on smaller parts of the dataset (the three first events, then 
just the first one), we see that the algorithm predicts anomalies at an almost
regular pace. More precisely, using the threshold that ensures the best
$\text{F1}_{pa}$ score results in a total of 4097 points predicted as anomalous,
of which 473 lie within an anomalous event and 3624 outside any anomalous event.
Knowing that the whole test dataset is about 5 days long, has one point per second,
and a total of 35 anomalous events, the algorithm, with such an output,
helps detect all 35 events while raising 3624 false alarms over 5 days. On average,
it raises an alarm every 110 seconds, making it, in our opinion, barely useful for
deployment. Such a behavior is generously rewarded by the \pa protocol.

\section{Discussion: Towards Better Practices for MVTS Anomaly Detection}
\label{sec:discussion}

\subsection{Evaluation Protocols and Metrics}
The \pa protocol has had its \emph{very illegitimate} hour of glory and created
what Wu and Keogh \cite{wu2021current} call ``illusion of progress'' in probably
quite a unique way in machine learning. We believe that this protocol should no
longer be used. Authors should resist the temptation of publishing high but misleading
scores using it, and reviewers should be aware of its flaws and advise authors to report
results using other protocols.

Evaluating algorithms for time series anomaly detection is not a trivial task.
The \pw protocol may be good as a straightforward way to compare and rank algorithms,
but other protocols and metrics, especially event-aware ones, should be used alongside
this protocol when possible. For datasets that contain anomalous events/episodes, at
least the number of detected events should be reported for each algorithm alongside \pw
metrics.

\subsection{Datasets and Experiment Design}
By sharing the raw versions of SWaT and Wadi datasets, the publishers apparently
wanted to give researchers the freedom to experiment with the data the way they want.
However, this has led to a divergence in the way these datasets are described and
used in the literature. We believe that it would be better that the data be stored
in a unique, accessible place using a format that anyone can start experimenting with
quickly. Ideally, part of the test data would be provided without labels so that
researchers upload the output of their algorithms to a third-party server for
evaluation and publication on a public leaderboard.

\subsection{Algorithms}
As mentioned in section \ref{sec:intro}, recent approaches for MVTS anomaly detection
have been marked by quite a few innovative ideas. However, based on the methodology
issues discussed in this paper, our position on this is the following: \emph{while
many of the proposed approaches are conceptually very appealing, we believe that the merits of
most of them are yet to be confirmed in the light of more appropriate evaluation protocols.}

Researchers should find a balance between the effort allocated to designing efficient
algorithms and that allocated to running sound experiments. Based on our experiments
and on the results shown by \cite{kim2022towards} using an \emph{untrained} model, it
is reasonable to assume that studies using the \pa protocol achieved high scores in a
relatively short time. Oftentimes in machine learning, an abnormally high score is achieved
with little effort either because the problem at hand is trivial, due to a data leakage
somewhere in the pipeline, or due to a bug in evaluation code. Admittedly, when the main
cause is a widely used evaluation protocol, it can be much harder to uncover, hence the
need to discuss evaluation protocols and their potential weaknesses and biases.


\bibliographystyle{splncs04}
\bibliography{main}

\begin{thebibliography}{10}
\providecommand{\url}[1]{\texttt{#1}}
\providecommand{\urlprefix}{URL }
\providecommand{\doi}[1]{https://doi.org/#1}

\bibitem{abdulaal2021practical}
Abdulaal, A., Liu, Z., Lancewicki, T.: Practical approach to asynchronous
  multivariate time series anomaly detection and localization. In: Proceedings
  of the 27th ACM SIGKDD Conference on Knowledge Discovery \& Data Mining. pp.
  2485--2494 (2021)

\bibitem{ahmed2017wadi}
Ahmed, C.M., Palleti, V.R., Mathur, A.P.: Wadi: a water distribution testbed
  for research in the design of secure cyber physical systems. In: Proceedings
  of the 3rd international workshop on cyber-physical systems for smart water
  networks. pp. 25--28 (2017)

\bibitem{audibert2020usad}
Audibert, J., Michiardi, P., Guyard, F., Marti, S., Zuluaga, M.A.: Usad:
  Unsupervised anomaly detection on multivariate time series. In: Proceedings
  of the 26th ACM SIGKDD International Conference on Knowledge Discovery \&
  Data Mining. pp. 3395--3404 (2020)

\bibitem{carmona2021neural}
Carmona, C.U., Aubet, F.X., Flunkert, V., Gasthaus, J.: Neural contextual
  anomaly detection for time series. Proceedings of the Thirty-First
  International Joint Conference on Artificial Intelligence, IJCAI-22  (2022)

\bibitem{chen2021daemon}
Chen, X., Deng, L., Huang, F., Zhang, C., Zhang, Z., Zhao, Y., Zheng, K.:
  Daemon: Unsupervised anomaly detection and interpretation for multivariate
  time series. In: 2021 IEEE 37th International Conference on Data Engineering
  (ICDE). pp. 2225--2230. IEEE (2021)

\bibitem{chen2021learning}
Chen, Z., Chen, D., Zhang, X., Yuan, Z., Cheng, X.: Learning graph structures
  with transformer for multivariate time series anomaly detection in iot. IEEE
  Internet of Things Journal  (2021)

\bibitem{deng2021graph}
Deng, A., Hooi, B.: Graph neural network-based anomaly detection in
  multivariate time series. In: Proceedings of the AAAI Conference on
  Artificial Intelligence. vol.~35, pp. 4027--4035 (2021)

\bibitem{garg2021evaluation}
Garg, A., Zhang, W., Samaran, J., Savitha, R., Foo, C.S.: An evaluation of
  anomaly detection and diagnosis in multivariate time series. IEEE
  Transactions on Neural Networks and Learning Systems  \textbf{33}(6),
  2508--2517 (2021)

\bibitem{han2022learning}
Han, S., Woo, S.S.: Learning sparse latent graph representations for anomaly
  detection in multivariate time series. In: Proceedings of the 28th ACM SIGKDD
  Conference on Knowledge Discovery and Data Mining. pp. 2977--2986 (2022)

\bibitem{hundman2018detecting}
Hundman, K., Constantinou, V., Laporte, C., Colwell, I., Soderstrom, T.:
  Detecting spacecraft anomalies using lstms and nonparametric dynamic
  thresholding. In: Proceedings of the 24th ACM SIGKDD international conference
  on knowledge discovery \& data mining. pp. 387--395 (2018)

\bibitem{kim2022towards}
Kim, S., Choi, K., Choi, H.S., Lee, B., Yoon, S.: Towards a rigorous evaluation
  of time-series anomaly detection. In: Proceedings of the AAAI Conference on
  Artificial Intelligence. pp. 7194--7201 (2022)

\bibitem{mathur2016swat}
Mathur, A.P., Tippenhauer, N.O.: Swat: A water treatment testbed for research
  and training on ics security. In: 2016 international workshop on
  cyber-physical systems for smart water networks (CySWater). pp. 31--36. IEEE
  (2016)

\bibitem{pan2022duma}
Pan, J., Ji, W., Zhong, B., Wang, P., Wang, X., Chen, J.: Duma: Dual mask for
  multivariate time series anomaly detection. IEEE Sensors Journal  (2022)

\bibitem{su2019robust}
Su, Y., Zhao, Y., Niu, C., Liu, R., Sun, W., Pei, D.: Robust anomaly detection
  for multivariate time series through stochastic recurrent neural network. In:
  Proceedings of the 25th ACM SIGKDD international conference on knowledge
  discovery \& data mining. pp. 2828--2837 (2019)

\bibitem{tuli2022tranad}
Tuli, S., Casale, G., Jennings, N.R.: {TranAD: Deep Transformer Networks for
  Anomaly Detection in Multivariate Time Series Data}. Proceedings of VLDB
  \textbf{15}(6),  1201--1214 (2022)

\bibitem{wu2021current}
Wu, R., Keogh, E.: Current time series anomaly detection benchmarks are flawed
  and are creating the illusion of progress. IEEE Transactions on Knowledge and
  Data Engineering  (2021)

\bibitem{xu2021anomaly}
Xu, J., Wu, H., Wang, J., Long, M.: Anomaly transformer: Time series anomaly
  detection with association discrepancy. The Tenth International Conference on
  Learning Representations  (2022)

\bibitem{zhang2019deep}
Zhang, C., Song, D., Chen, Y., Feng, X., Lumezanu, C., Cheng, W., Ni, J., Zong,
  B., Chen, H., Chawla, N.V.: A deep neural network for unsupervised anomaly
  detection and diagnosis in multivariate time series data. In: Proceedings of
  the AAAI conference on artificial intelligence. vol.~33, pp. 1409--1416
  (2019)

\bibitem{zhang22grelen}
Zhang, W., Zhang, C., Tsung, F.: Grelen: Multivariate time series anomaly
  detection from the perspective of graph relational learning. In: Proceedings
  of the Thirty-First International Joint Conference on Artificial Intelligence
  (IJCAI) (2022)

\bibitem{zong2018deep}
Zong, B., Song, Q., Min, M.R., Cheng, W., Lumezanu, C., Cho, D., Chen, H.: Deep
  autoencoding gaussian mixture model for unsupervised anomaly detection. In:
  International conference on learning representations (2018)

\end{thebibliography}
\end{document}